\documentclass[preprint,12pt]{elsarticle}

\usepackage{amssymb}
 \usepackage{amsmath}
\usepackage{hyperref}
\usepackage{multirow}
\usepackage{xcolor,colortbl}

\newcommand{\best}[1]{\cellcolor{gray}\textcolor{white}{\textbf{{#1}}}}
\newcommand{\bestcnn}[1]{\cellcolor{black}\textcolor{white}{\textbf{{#1}}}}

\journal{GECCO 2024 (accepted)}

\begin{document}

\begin{frontmatter}



\title{Creating Ensembles of Classifiers through UMDA for Aerial Scene Classification \\\textcolor{red}{(extended version)}}


\author{Fabio A. Faria, Luiz H. Buris, Luis A. M. Pereira, F\'{a}bio A. M. Cappabianco, }

\affiliation{organization={Institute of Science and Technology -- Universidade Federal de Sao Paulo},%Department and Organization
            addressline={Avenida Cesare Mansueto Giulio Lattes, 1201,  Eugenio de Mello}, 
            city={Sao Jose dos Campos},
            postcode={CEP 12247014}, 
            state={Sao Paulo},
            country={Brazil}}

\begin{abstract}
Aerial scene classification, which aims to semantically label remote sensing images in a set of predefined classes (e.g., agricultural, beach, and harbor), is a very challenging task in remote sensing due to high intra-class variability and the different scales and orientations of the objects present in the dataset images. In remote sensing area, the use of deep learning architectures as an alternative solution is also a reality for scene classification tasks. Generally, these architectures are used to perform the traditional image classification task. However, another less used way to classify remote sensing image might be the one that uses deep metric learning (DML) approaches. In this sense, this work proposes to employ six DML approaches for aerial scene classification tasks, analysing their behave with four different pre-trained architectures (convolutional neural networks -- CNNs) as well as combining them through evolutionary computation algorithms. In performed experiments, it is possible to observe than DML approaches can achieve the best classification results when compared to traditional pre-trained CNNs for three well-known remote sensing aerial scene datasets (AID, UCMerced, and RESISCS). In addition, the UMDA algorithm proved to be a promising strategy to combine DML approaches when there is diversity among them, managing to improve at least 5.6\% of mean accuracy in the classification results using almost 50\% of the available classifiers for the construction of the final ensemble of classifiers.
\end{abstract}


\begin{highlights}
\item First work in the literature comparing performance between traditional pre-trained CNNs and Deep Metric Learning (DML) approaches for aerial scene classification tasks;
\item A study of the existing diversity among DML-based classifiers for aerial scene classification tasks;
\item Creating ensembles of DML-based classifiers based on Univariate Marginal Distribution Algorithm (UMDA) for aerial scene classification tasks;
\end{highlights}


\end{frontmatter}



\section{Introduction}
In recent years, advances in remote sensing have made classification tasks more difficult, increasing the level of abstraction from pixels to objects and, finally, scenes~\cite{Cheng2020RemoteSI,CastelluccioPSV15}. Traditional pixel-based approaches use spectral responses (e.g., RGB channels and NDVI) to determine their categories. In Geographic Object-based Image Analysis (GEOBIA) or object-based methods, instead of assigning categories to each pixel in the image, these methods aim to identify specific objects of interest, such as streets, lakes, and buildings, to enable scene classification. With the emergence of many aerial scene datasets, the scene classification task follows the semantics of the entire image, i.e., methods classify all pixels of an image with the same category~\cite{Chew2018}.
 
Aerial scene classification is a challenging remote sensing task due to high intra-class variability and the {different scales and orientations of objects} in the dataset images~\cite{CastelluccioPSV15}. This task has {applications} in military and civil areas, such as natural disaster monitoring, weapon guidance, and traffic supervision~\cite{Yunlong2018}. There are several extensive studies devoted to aerial scene classification, and proposed methods lay into (1) low-level, (2) mid-level, and (3) deep-level categories~\cite{Yunlong2018}.
Low-level methods utilize local descriptors such as Scale Invariant Feature Transform (SIFT)~\cite{sift2004}, color histogram (CH)~\cite{gch}, and Local Binary Patterns (LBP)~\cite{lbp1990}. Mid-level methods apply local descriptor encoding (e.g., SIFT) to create a representation with more semantic information, such as Bag-of-Visual-Words (BoVW)~\cite{Avila2011bossa}.
Deep-level methods adopt deep learning architectures (e.g., VGG16~\cite{vgg16} and ResNet~\cite{he2016deep}) to extract more discriminative visual properties with semantic-level information from the images.  
 
Due to the ability to robustly extract semantic-level information, convolutional neural networks (CNNs) have dominated research solutions in computer vision and machine learning for the last ten years in different applications (e.g., action recognition~\cite{YAO2018}, biometric recognition~\cite{Sundararajan:2018}, and medical image analysis~\cite{Ker2018}).
Several researchers participated in research challenges/competitions (e.g., IARPA~\cite{fmow2018} and GRSS~\cite{GRSS2018}) to develop robust techniques capable of correctly classifying as many test samples as possible. Among these solutions, CNN architectures are the most influential ones.  

\begin{figure}[ht!]
    \centering
    \includegraphics[scale=.6]{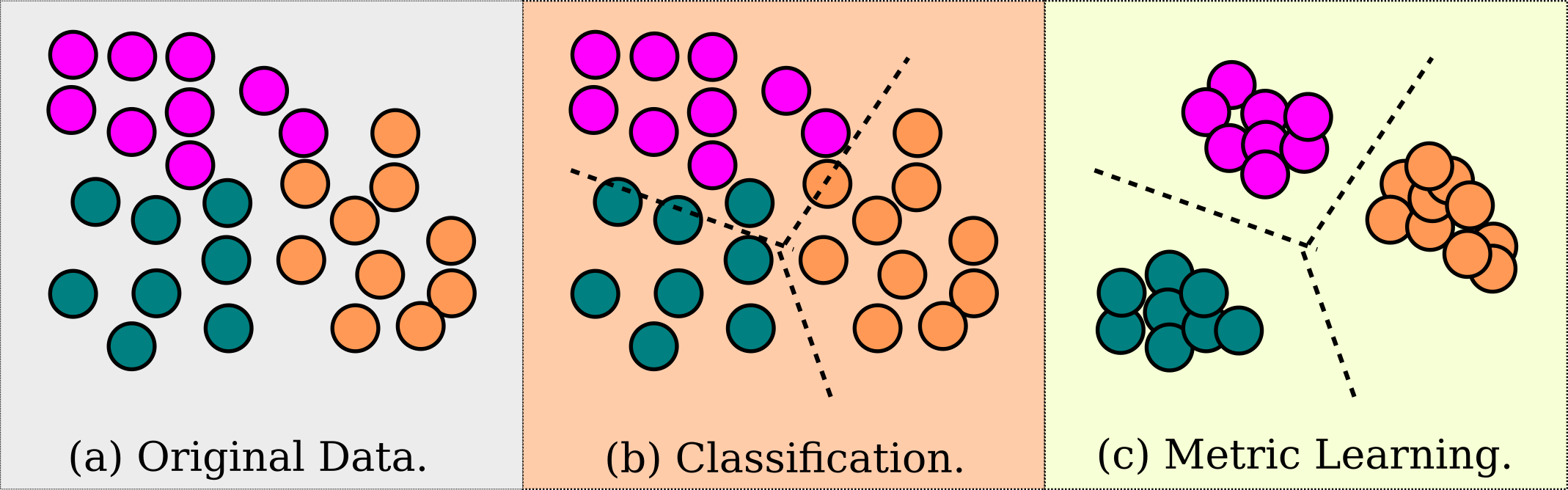}
    \caption{Difference between traditional classification and metric learning approaches.}
    \label{fig:comparison}
\end{figure}

Image classification tasks may use traditional machine learning models (e.g. decision trees~\cite{dt-survey}, linear regression models~\cite{Bishop_2006}, support vector machines~\cite{svm}, and neural networks~\cite{wolpert_1992}), which learn a decision boundary (Fig.\ref{fig:comparison}-b) base on original data of different classes (Fig.\ref{fig:comparison}-a).

Among the many works proposed in the literature, it is worth highlighting Penatti et al.~\cite{Penatti_2015_CVPR_Workshops}, which they proposed a comparative evaluation of global descriptors, BoVW descriptors, and CNNs in aerial and remote sensing domain as well as a correlation analysis among different CNN and among different descriptors. Lin et al.~\cite{lin_grsl2017} introduced an unsupervised model called multiple-layer feature-matching generative adversarial networks (MARTA GANs) to learn a representation of unlabeled data. Minetto et al.~\cite{hydra2018} proposed a solution called Hydra, which creates ensembles of CNNs based on ResNet and DenseNet models. Basically, Hydra uses a coarsely optimized CNN to create the Hydra's body. Then, the weights are fine-tuned multiple times to build multiples CNNs using data augmentation to represent the Hydra's heads. Yu et al.~\cite{yu_tgrs2019} proposed a novel unsupervised deep feature learning method called Attention generative adversarial networks (Attention GANs), which it enhances the representation power of the discriminator through the use of attention module~\cite{vaswani_nips2017} and context aggregation-based feature fusion strategy. Len et al.~\cite{he_tnnls2020} embed two novel modules into the traditional CNN models (skip connections and covariance pooling) to achieve more representative feature learning when dealing with scene classification problems.

Another way to perform classification tasks, very little explored in the field of remote sensing, is to employ  metric learning approaches. These approaches build a new projected/transformed space (transformation matrix) from the original data (Fig.\ref{fig:comparison}-a), mapping same-class data closer together and distinct-class data further away (Fig.\ref{fig:comparison}-c) ~\cite{kulis2013metric}. Furthermore the combination of metric learning approaches and deep learning architectures has been extensively explored in the literature, giving rise to a new concept known as deep metric learning (DML). DML employs deep neural networks to learn the transformed space (metric), moving distinct-class data away and approximating same-class data in the target application~\cite{hav4ik2021deepmetriclearning}.

DML approaches may belong to three different categories: (1) cost/loss function (Contrastive~\cite{Contrastive_cvpr2006}, Triplet~\cite{triplet_cnn_cvpr2015}, and ProxyAnchor~\cite{ProxyAnchor_cvpr2020}), (2) neighborhood (NCA~\cite{goldberger2004neighbourhood}, proxy-NCA~\cite{movshovitz2017no}, and NNGK~\cite{NNGK_ICIPP2018}), and (3) those from custom layers (SoftTriple~\cite{SoftTriple_iccv2019} and SupCon~\cite{SupCon_nips2020}). As these approaches use deep learning architectures, they must overcome the same limitations (e.g., overfitting and vanish gradient) to achieve satisfactory results in the target applications. 

In terms of remote sensing scene classification, to the best of our knowledge, there is only one work in the literature, which deep metric learning approach has been applied into this context. Cheng et al.~\cite{cheng_2018} proposed the discriminative CNNs (D-CNN), which it combined cross-entropy function and metric learning regularization term on the final objective function to improve the classification results. Nevertheless, many other works have adopted DML  approaches for different remote sensing tasks such as chance detection~\cite{zhang_rs2022}, remote sensing image retrieval~\cite{yun_rs2020,zhao_grsl2022}, hyperspectral image classification~\cite{dong_rs2021}, and image classification with noisy annotations~\cite{kang_igarss_2021}.

In this sense, observing the lack of works in the literature that adopt DML approaches for scene classification tasks, the main contributions of this paper are three-fold:
\begin{itemize}
    \item A comparative analysis among six DML approaches and four deep-learning architectures (DLA), resulting in twenty-four different classifiers for aerial scene classification tasks;
    \item An analysis of the existing diversity among all available DML-based classifiers for aerial scene classification tasks;
    \item An approach of DML-based classifier fusion based on UMDA algorithm to improve the classification results in aerial scene classification tasks.  
\end{itemize}

\section{Background}

This section presents the essential concepts for a better understanding of this work.

\subsection{Deep Metric Learning (DML)}

Deep metric learning (DML) is the junction of two concepts: deep learning and metric learning. In DML, a deep neural network  automatically learns data feature space, reducing same-class example distances and increasing distinct-class example distances. These are the deep metric learning leverage categories: (1) cost/loss function (Contrastive~\cite{Contrastive_cvpr2006}, Triplet~\cite{triplet_cnn_cvpr2015}, and ProxyAnchor~\cite{ProxyAnchor_cvpr2020}), (2) neighborhood (NCA~\cite{goldberger2004neighbourhood} and NNGK~\cite{NNGK_ICIPP2018}), and (3) custom layers (SoftTriple~\cite{SoftTriple_iccv2019} and SupCon~\cite{SupCon_nips2020}).

\subsubsection{{Contrastive}}

%\faf{trocar os ponto $x_1$ por $x_p;x_n$ }

%Contrastive function~\cite{hav4ik2021deepmetriclearning} is a classical loss function for metric learning and it was used to retrieve similar images in queries for face verification applications. It consists of gathering similar examples, i.e., examples of the same class while different examples of different classes must be further away in the observed space of the target set. Thus, contrasting between observed examples of positive pairs and negative pairs, the examples of positive pairs will be closer to each other, while the negative pairs are more distant.

A contrastive function~\cite{hav4ik2021deepmetriclearning} is a classical metric learning loss function used to retrieve similar images in face verification applications. It gathers same-class (positive) examples, while distinct-class (negative) examples must lay further away in the observed target set space. 
%Thus, as we contrast positive and negative pairs of examples, positive pairs stay closer to each other, while negative ones are more distant.

%Considering Equation~\ref{Eq:losscontrast}, two points $x_p$ and $x_n$ which are examples of the dataset within the batch, $y_p$ and $y_n$ are their corresponding classes, to calculate the examples of equal class (${\cal L}_{y_p=y_p}$), calculate the distance $D^2_{f^\theta}(x_p,x_p)$ between the feature vectors of the points considered and provided by the function of neural network $f^{\theta}$ and the distance between the examples is desired to be as small as possible, indicating that the examples belong to the same class. For examples of different classes (${\cal L}_{y_p \neq y_n}$), the calculation performed is $max(0, m - D^2_{f^\theta}(x_p,x_n))$ and it should ensure that the distance is as large as possible, so that the loss is always minimised. To prevent the neural network $f^\theta$ from learning to ``cheat'', all $x_{i}$ are mapped to the same point making the distances between any examples equal to zero, a margin value $m$ is adopted.

The contrastive function in Equation~\ref{Eq:losscontrast} contains a same-class and a distinct-class term, ${\cal L}_{eq}$ and ${\cal L}_{df}$, respectively.  Let $p$ and $n$ be any pair of examples, where $x_p$ and $x_n$ are their feature vectors and $y_p$ and $y_n$ are their corresponding classes. If $y_p = y_n$, ${\cal L}_{eq}$ calculates the distance $D^2_{f^\theta}(x_p,x_n)$ between their feature vectors, based on the neural network function $f^{\theta}$. Otherwise, ${\cal L}_{df}$ computes the difference between a margin distance value $m$ and the distance between their features, with a minimum distance of zero. Adopting a margin distance value prevents the neural network $f$ from mapping all same-class examples to the same point, resulting in a zero distance among them.
\vspace{-0.25cm}
\begin{eqnarray}
\label{Eq:losscontrast}
%{\cal L}_{con} & = & {\cal L}_{eq} + {\cal L}_{df} \\
{\cal L}_{eq} & = & {\cal L}_{y_p=y_n}D^2_{f^\theta}(x_p,x_n) \nonumber \\
{\cal L}_{df} & = & {\cal L}_{y_p\neq y_n} max(0, m - D^2_{f^\theta}(x_p,x_n)) \nonumber \\
{\cal L}_{con} & = & {\cal L}_{eq} + {\cal L}_{df}
\end{eqnarray}

\subsubsection{{Triplet}}

The Triplet loss function~\cite{hav4ik2021deepmetriclearning} widely appears in learning metrics for facial recognition. Let $a$, $p$, and $n$ be an anchor, a positive, and a negative example, respectively. $a$ and $p$ belong to the same class, i.e. $y_a = y_p$, and $n$ belongs to a distinct class $y_n$. Each example has its feature vector $x_a$, $x_p$, and $x_n$. The Triplet loss function minimizes the distance between $x_a$ and $x_p$ while maximizing the distance between $x_a$ and $x_n$, observing a marginal distance of $m$. Equation~\ref{Eq:lossTriplet} describes the triplet function.
\vspace{-0.25cm}
\begin{eqnarray}
{\cal L}_{pos} & = & D^2_{f^\theta}(x_a,x_p)\nonumber \\
{\cal L}_{neg} & = & D^2_{f^\theta}(x_a,x_n) \nonumber \\
{\cal L}_{triplet} & = & max(0,{\cal L}_{pos} - {\cal L}_{neg} + m) \label{Eq:lossTriplet} 
\end{eqnarray}

\subsubsection{{Nearest Neighbour Gaussian Kernels (NNGK)}}
\label{sec:nngk}

NNGK~\cite{NNGK_ICIPP2018} is an excellent approach for image classification based on Neighbourhood Components Analysis (NCA)~\cite{goldberger2004neighbourhood}. NNGK calculates the Euclidean distance between the training data of resource $x$ and its center $c$ in the training set. This way, computing the Kernel curvature width ({bandwidth}) constrains the calculation of the distance of the kernel function, the value of the width of the kernel, and the same value for all centers present in the problem space.

Equation~\ref{eq:kg} defines the Gaussian kernel $f$:

\begin{equation}
    f(x,c)=\exp\left(\frac{-\|x-c\|^2}{2\phi^2}\right),
\label{eq:kg}
\end{equation}

A classifier is the weighted sum of the kernel function distance between an example's feature vector ({embeddings}) and its respective center. The classification of an example $x$ results from transporting information across the network, generating a high-dimensional feature vector in the same space as the center designed by kernel~\cite{NNGK_ICIPP2018}. To calculate the kernel function, we train the high-dimensional intermediate layer before the loss function or classifier as the feature embedding layer.

Equation~\ref{Eq:kclassification} presents the probability of an example $x$ belonging to a label class. The end-to-end network is responsible for learning the weights $w_i$ to each center $i$. $f(x,c)$ is the kernel function and $\cal{M}$ is the number of training examples. Note that each example contributes to a single training set center.

\begin{equation}
    Pr(x \in class\ Q)= \frac{\sum_{i \in Q}{w_{i}f(x,c_{i})}} {\sum_{j=1}^{\cal M}{w_{j}f(x,c_{j})}}
    \label{Eq:kclassification}
\end{equation}

Equation~\ref{Eq:nngk} defines the NNGK classifier. $\mathcal{N}$ is the set of nearest neighbors for example $x$ and $i$ $\in Q \cap \mathcal{N}$ is the set of labeled nearest neighbors $R$. %Again,
Training set examples exclude their own center from their list of nearest neighbors.

\begin{equation}
     Pr(x \in class\ R)= \frac{\sum_{i \in Q \cap \mathcal{N}}{w_{i}f(x,c_{i})}} {\sum_{j \in \mathcal{N}}{w_{i}f(x,c_{j})}}
    \label{Eq:nngk}
\end{equation}

Equation~\ref{Eq:logNeg} shows the NNGK loss function as the sum of the negative logarithmic probability of the probabilities of $x$ belonging to the correct label $R$.

\begin{equation}
     {\cal L}_{NNGK} = -\ln{(Pr(x\ \in \ class\ R))}
    \label{Eq:logNeg}
\end{equation}

\subsubsection{{ProxyAnchor}}

The ProxyAnchor loss function proposed in~\cite{ProxyAnchor_cvpr2020} consists of taking each proxy as an anchor and associating it with all the data in a batch so that it pulls or pushes data in different degrees of force according to their relative hardness (relationships between data during training). A proxy is a subset representative of training data estimated as part of the network parameters.

Equation~\ref{Eq:proxyanchor} formulates the ProxyAnchor function, where $m > 0$ is the margin and $\alpha$ is the scale factor. $P$, $P^+$, and $P^-$ indicate the sets of all proxies, the positive, and the negative proxy datasets within the batch, respectively. $P$ is a batch of the feature vectors $X^p$ divided into datasets $X^+_p$ and $X^-_p $. $X^+_p$ and $X^-_p = X^p - X^+_p$ are the sets of positive and negative trait vectors from lot $p$, respectively.

\begin{equation}
\begin{aligned}
{\cal L}_{PA}(X) = {} & \frac{1}{|P^+|} \sum_{p \epsilon P^+} log \left( 1+ \sum_{x \epsilon X_p^+} e^{-\alpha(s(x,p)-m)} \right)   \\  & + \frac{1}{|P|} \sum_{p \epsilon P^-} log \left(  1+ \sum_{x \epsilon X_p^-} e^{\alpha(s(x,p)+m)} \right)
\end{aligned}
\label{Eq:proxyanchor}
\end{equation}

\subsubsection{{SoftTriple}}
SoftTriple loss~\cite{qian2019softtriple} function appears as an alternative to the conventional Softmax loss function, equivalent to the smoothed Triplet loss function where each label has a single weight vector. Instead of the traditional Softmax loss function with only one weight vector per label, SoftTriple uses a Softmax with several weight vectors in the fully connected layer FC layer.

Therefore, it presents much more satisfactory results in real-world applications since a label contains several local representatives. Adding a fully connected layer to the top of a classification model increases the number of center or weight vectors between the feature space layer and the softmax operator. In this sense, different operators generate similarities, calculating the distribution of the output probability vectors for distinct labels.

Equation~\ref{Eq:softtriple2} defines the softmax smoothed similarity $S_{i,y_i}$ of an input data $x_i$ with label $y_i$, using the center weight vectors $W^k_{y_i}$. Each label adopts a $\gamma$ acting as an entropy regularizer for the distribution to prevent ties among centers.

\begin{equation}
S_{i,y_i} = \sum_{k} \frac{exp(\frac{1}{\gamma} x_i^{T} W^{k}_{y_{i}})}{\sum_{k} exp(\frac{1}{\gamma} x^T_i W^{k}_{y_{i}})} x_i^T W^{k}_{y_{i}}
\label{Eq:softtriple2}
\end{equation}

Equation~\ref{Eq:softtriple1} defines the SoftTriple loss function with the smoothed similarity, where $\lambda$ is a scale factor over the smoothed similarity of the softmax and $m$ is a predefined margin.
%\vspace{-0.15cm}

\begin{equation}
{\cal L}_{ST}(x_i) = -log  \frac{exp(\lambda(S_{i,y_i} - m))}{exp(\lambda(S_{i,y_i} - m)) + \sum_{j \neq y_i} exp(\lambda S_{i,j}) } 
\label{Eq:softtriple1}
\end{equation}

\subsubsection{{Supervised Contrastive (SupCon)}}

SupCon is a custom layer-based metric learning technique proposed in~\cite{khosla2020supervised}. It applies a data augmentation operation over an input dataset, obtaining two copies of the batch of images. They are later passed through two deep neural networks to generate their feature vectors (embeddings). In the inference stage, the contrastive loss is computed using only one of the networks with it weights frozen and then a linear classification model is used to minimize the cross-entropy loss function. Equation~\ref{Eq:supConLoss} defines the SupCon loss function adopted in this work, where, $x_a$ is the anchor, $x_p$ is the positive example and $\tau$ is a parameter called ``temperature" to control the training optimization. The loss is the positive logarithm in the equation. To minimize the loss, the inner product between the anchor $x_a$ and the positive example $x_p$ must be low, and the one between the anchor $x_a$ and the negative example $x_n$ must be high.

\begin{equation}
\mathcal{L}_{SC} = \sum_{i \epsilon l} \frac{-1}{|P(i)|} \sum_{p \epsilon P(i)} log \frac{exp(x_a \cdot x_p/ \tau)}{\sum_{a \epsilon A(i)}  exp(x_a \cdot x_n / \tau)}
\label{Eq:supConLoss}
\end{equation}

\subsection{Ensembles of Classifiers and Diversity Concept}
\label{sec:diversity_measure}

Ensembles of classifiers consists of a set of different machine learning models (base classifiers) and a fusion strategy for combining classifier outputs. Ensemble of classifiers aims to create a more generalizable model, combining complementary visual information and improving the effectiveness of results in a single target task (image classification)~\cite{roli_mcs_2002,Polikar_2012,faria_prl_2014}.
%https://link.springer.com/referenceworkentry/10.1007/978-0-387-73003-5_148

In ensembles of classifiers, an essential {point} to yield good classification performance is called diversity, which measures the degree of agreement/disagreement among classifiers within the ensemble of classifiers through a relationship matrix~\cite{Kuncheva_2004, Kuncheva2009}.

Let $\mathcal{M}$ be a matrix containing the relationship among a pair of classifiers with the percentage of concordance 
that compute the percentage of \textit{hit} and \textit{miss} for two exemplifying classifiers $c_i$ and $c_j$.
The value $a$ is the percentage of examples that both classifiers $c_i$ and $c_j$ classified correctly in a validation set.
Values $b$ and $c$ are the percentage of examples that $c_j$ hit and
 $c_i$ missed and vice-versa. The value $d$ is the percentage of examples that both classifiers missed.
   
 \begin{table} [htp!]% 
 \centering
 \caption{Relationship matrix $\mathcal{M}$ among pairs of classifiers $c_i$ and $c_j$.}
%  \resizebox{4.5cm}{!} {
 \begin{tabular}{|l|c|c|} \hline
 & Hit $c_i$ & Miss $c_i$ \\ \hline \hline
 Hit $c_j$&  $a$            & $b$ \\ \hline
 Miss   $c_j$&$c$            & $d$ \\ \hline
 \end{tabular}
%  }
 \label{tab:diversidade}
 \end{table}

\textit{Among many diversity measures studied by Kuncheva et al.~\cite{ Kuncheva2009}, we can define the Correlation Coefficient ($\rho$)} measure as follows:

\begin{equation}
\label{eq:cor}
\rho(c_i,c_j) =\frac{ad-bc}{\sqrt{(a+b)(c+d)(a+c)(b+d)}},
\end{equation}

Diversity is higher if the score of the \textit{Correlation Coefficient $p$} is lower between a pair of classifiers $c_i$ and $c_j$.

\subsection{Univariate Marginal Distribution Algorithm (UMDA)}

UMDA~\cite{muhlenbein1996recombination} is known as one of the simplest Estimation of Distribution Algorithms~\cite{HAUSCHILD2011111} (EDAs). To optimize a pseudo-Boolean function $f: \{0,1\}^n \rightarrow \mathbb{R}$ where an individual is a bit-string (each gene is $0$ or $1$), the algorithm follows an iterative process: 1) independently and identically sampling a population of $\lambda$ individuals (solutions) from the current probabilistic model; 2) evaluating the solutions; 3) updating the model from the fittest $\mu$ solutions. Each sampling-update cycle is called a generation or iteration. In each iteration, the probabilistic model in the generation $t \in \mathbb{N}$ is represented as a vector $p_t=(p_t(1),...,p_t(n))\in [0,1]^n$, where each component (or marginal) $p_t(i)\in[0,1]$ to $i\in[n]$, and $t \in \mathbb{N}$ is the probability of sampling the number one in the $i^{th}$ position of an individual in generation $t$. Each individual $x = (x_1, ..., x_n) \in \{0, 1\}^n$ is therefore sampled from the joint probability.
 \begin{equation}
    Pr(x | p_t) = \prod_{i=1}^{n}p_t(i)^{x_i}(1-p_t(i))^{(1-{x_i})}.
    \label{equacaoPr}
\end{equation}

In literature, UMDA is superior in terms of speed, memory consumption, and accuracy of solutions than genetic algorithms~\cite{Hashemi2011}. In addition, this algorithm has been successfully used for creating effective ensembles of classifiers through stochastic selection/pruning of the available classifiers and achieved excellent results in scene classification tasks~\cite{Ferreira_icpr2020}.

\section{Experimental Methodology}
This section presents the experimental methodology adopted in this work for the aerial scene classification task.

\subsection{Deep Metric Learning (DML)}

Concerning the deep metric learning approaches,  Contrastive, ProxyAnchor, SoftTriple, SupCon, and Triplet, we used the source codes available at PyTorch Metric Learning lib~\footnote{\tiny{\url{https://kevinmusgrave.github.io/pytorch-metric-learning/}} (As of September 2022)} with their default parameters. Furthermore, we used source code provided by the authors of NNGK\footnote{\tiny{\url{https://github.com/LukeDitria/OpenGAN}} (As of September 2022)} approach. Two center values $|\mathcal{C}|=\{100,200\}$ and two different values for $\sigma=\{5,10\}$ are the parameters of the NNGK approach.

\subsection{Deep Learning Architectures (DLA)}

The pre-trained CNN architectures (PT-CNN) in this work are two deep residual networks (ResNet18 and ResNet50~\cite{he2016deep}) and two VGG-based neural networks (VGG16 and VGG19~\cite{vgg16}). As cross-entropy loss is the most widely used as loss
function to train CNNs, thus it has been adopted in these experiments.  Table~\ref{tab:cnns} shows more details about those deep learning architectures.

\begin{table}[ht!]
    \centering
    \caption{Original deep learning architecture (DLA) configuration.}
    \resizebox{11cm}{!}{ 
    \begin{tabular}{|c|ccc|} \hline
         \textbf{DLA} & \textbf{Weight Layers}  & \textbf{Parameters} & \textbf{Feature Vector Size}\\\hline
         ResNet18 & 18 & 11.0M  & 512 \\ 
         ResNet50 & 50 & 25.6M & 2048 \\
         VGG16    & 16 & 138.0M& 4096 \\
         VGG19    & 19 & 144.0M & 4096 \\\hline
    \end{tabular}
    }
    \label{tab:cnns}
\end{table}

We trained all four pre-trained deep learning architectures by $200$ epochs with a batch size of $32$ images implemented on the PyTorch library~\cite{pytorch_NEURIPS2019_9015} as the backbone for all deep metric learning approaches. We ran all the experiments on an NVIDIA GeForce GTX Titan V 12GB.

\subsection{{Aerial Scene Datasets}}

To evaluate the classification results of deep metric learning approaches (DML) and the impact of deep learning architectures (DLA) in aerial scene classification tasks, we adopt three well-known aerial scene datasets (AID~\cite{xia2017aid}, UCMerced~\cite{yang2010bag}, and RESISC45~\cite{cheng2017remote}) in this work. All datasets use a $5$-fold cross-validation protocol separated into three sets: training ($20\%$), validation ($20\%$), and test ($60\%$) sets.  Therefore, the training set is the input for all classifiers composed of DML$+$DLA, which will classify new images from the test set. The validation set only serves as input for the classifier fusion strategy (UMDA~\cite{muhlenbein1996recombination}), which selects the most suitable classifiers applied to the new images from the test set. %\textcolor{red}{descrever abaixo:}

\begin{itemize}
\item  \textbf{Aerial Image Dataset (AID)~\cite{xia2017aid}\footnote{\tiny{\url{https://captain-whu.github.io/AID/}}}:} 10000 images, 30 different scene classes and about 200 to 400 samples of size $600 \times 600$ in each class. \vspace{-0.25cm}
\item  \textbf{UCMerced~\cite{yang2010bag}\footnote{\tiny{\url{http://weegee.vision.ucmerced.edu/datasets/landuse.html}}}:}  2100 images, 21 scene classes and 100 samples of size $256 \times 256$ in each class. \vspace{-0.25cm}
\item  \textbf{RESISC45~\cite{cheng2017remote}\footnote{\tiny{\url{https://www.tensorflow.org/datasets/catalog/resisc45}}}:} 31,500 images, 45 scene classes with 700 samples of size $256 \times 256$ in each class.
\end{itemize} 

\begin{figure*}[ht!]
    \centering
\begin{tabular}{ccccccc}

\includegraphics[scale=.17]{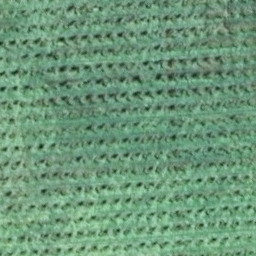}&
\includegraphics[scale=.17]{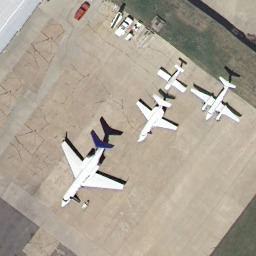}&
\includegraphics[scale=.17]{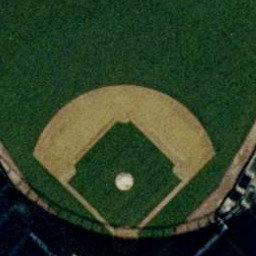}&
\includegraphics[scale=.17]{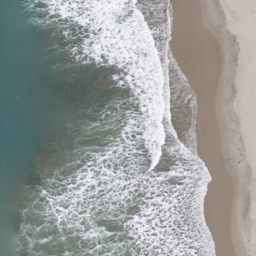}&
\includegraphics[scale=.17]{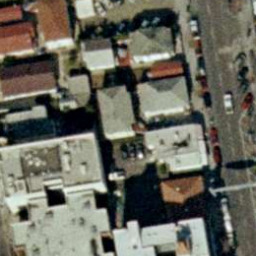}&
\includegraphics[scale=.17]{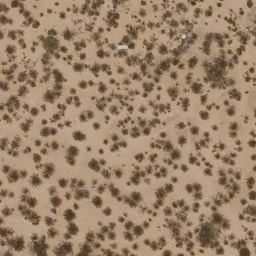}&
\includegraphics[scale=.17]{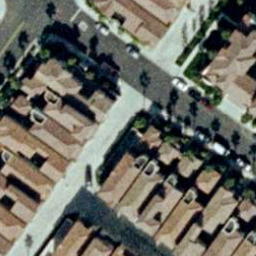}\\ 
\multirow{2}{*}{\tiny{Agricultural}}&\multirow{2}{*}{\tiny{Airplane}}&\tiny{Baseball}&\multirow{2}{*}{\tiny{Beach}}&\multirow{2}{*}{\tiny{Building}}&\multirow{2}{*}{\tiny{Chaparral}}& \tiny{Dense}  \\ 
&&\tiny{Diamond} &&&& \tiny{Residential}\\
\includegraphics[scale=.17]{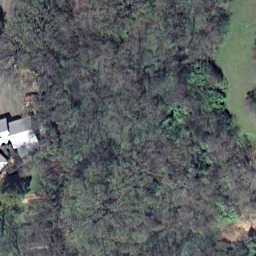}&
\includegraphics[scale=.17]{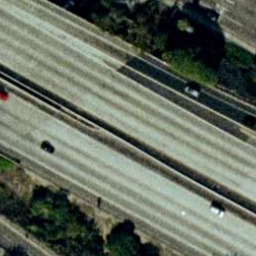}&
\includegraphics[scale=.17]{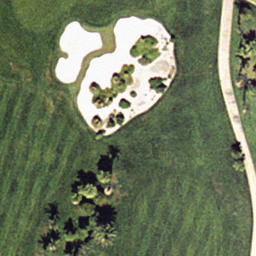}&
\includegraphics[scale=.17]{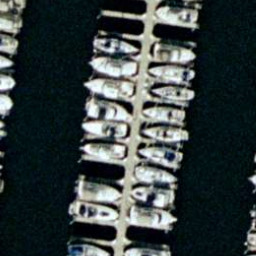}&
\includegraphics[scale=.17]{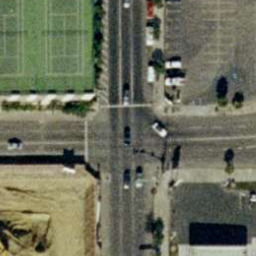}&
\includegraphics[scale=.17]{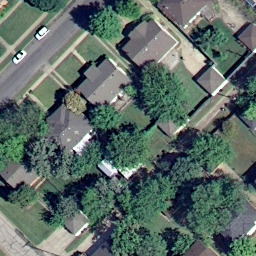}&
\includegraphics[scale=.17]{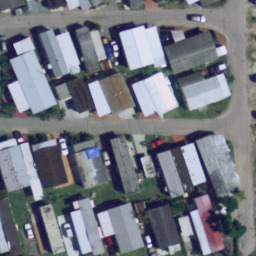}\\ 
\multirow{2}{*}{\tiny{Forest}}&\multirow{2}{*}{\tiny{Freeway}}&\tiny{Golf} &\multirow{2}{*}{\tiny{Harbor}}&\multirow{2}{*}{\tiny{Intersection}}&\tiny{Medium}&\tiny{Mobile} \\ 
&&\tiny{Course}&&&\tiny{Residential}& \tiny{Homepark}\\
\includegraphics[scale=.17]{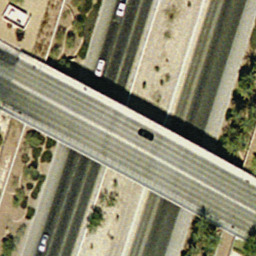}&
\includegraphics[scale=.17]{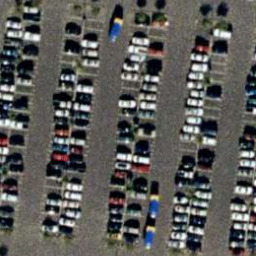}&
\includegraphics[scale=.17]{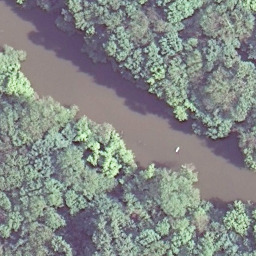}&
\includegraphics[scale=.17]{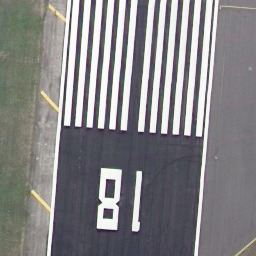}&
\includegraphics[scale=.17]{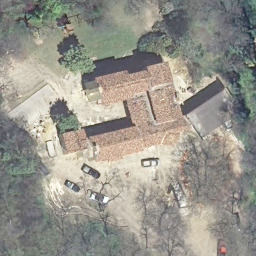}&
\includegraphics[scale=.17]{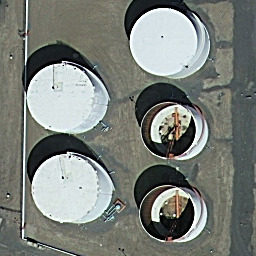}&
\includegraphics[scale=.17]{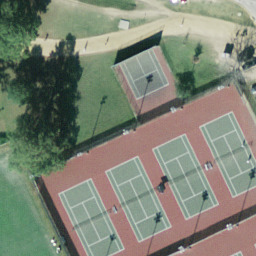}\\ 
\multirow{2}{*}{\tiny{Overpass}}&\multirow{2}{*}{\tiny{Parkinglot}}&\multirow{2}{*}{\tiny{River}}&\multirow{2}{*}{\tiny{Runway}}&\tiny{Sparse}&\tiny{Storage}& \tiny{Tennis} \\ 
&&&&\tiny{Residential}& \tiny{Tanks}& \tiny{Court}\\
\end{tabular}  
    \caption{Examples of aerial scenes from UCMerced dataset used in this work.}
    \label{fig:dataset}
\end{figure*}

Figure~\ref{fig:dataset} shows some examples for each one of $21$ classes existing in the UCMerced dataset.

\section{Experiments and Discussion}
This section shows and discusses all experiments in this work.

\subsection{Finding the Best Classifiers (DML$+$DLA)}
%\textcolor{red}{Parei aqui!}

Table~\ref{tab:overall} shows the classification results of six deep metric learning approaches existing in the literature (Contrastive, ProxyAnchor, SoftTriple, SupCon Triplet, and NNGK), four well-known deep learning architectures (ResNet18, ResNet50, VGG16 and VGG19~\cite{alzubaidi2021review}) over three remote sensing image datasets (AID, UCMerced and RESISC45). The goal is to find a combination between the best tuple composed of a deep metric learning (DML) approach and a deep learning architecture (DLA) for the different image datasets, as well as to identify which DLA achieves the best classification results for this target application. Furthermore, the traditional architectures of pre-trained CNNs with no DML approach (PT-CNN) were adopted as baseline approaches as well.

As the best combinations (in gray) are SupCon$+$ResNet18 ($87.37\%$) for AID dataset, ProxyAnchor$+$VGG19 ($77.70\%$) for UCMerced dataset, and NNGK$+$Resnet50 ($91.60\%$) for RESISC45 dataset.  
In addition, DLA can directly influence the performance of DML in different datasets. DML approaches may get better or worse depending on the associated DLA for each dataset.

For instance, in the AID and UCMerced datasets, ProxyAnchor and SoftTriple approaches achieved good classification results regardless of the DLA used in these experiments. SupCon approach achieved excellent classification results with ResNet architecture, however it did not the same performace when employed VGG architecture. 

In RESISC45 dataset, the NNGK approach achieved good classification results using any DLA. However, Contrastive, SoftTriple, SupCon, and Triplet suffered an enormous drop in performance (more than $20\%$ among the best DML approach -- NNGK) while using ResNet18 architecture.

In terms of the final ranking positions (\textit{$@$R}), Softtriple is the most stable DML among all the compared approaches for the AID dataset, ProxyAnchor for the UCMerced dataset, and NNGK for the RESISC45 dataset. In addition, comparing the performance of all DML approaches and considering their positions in the final ranking (Final \textit{$@$R}), the ProxyAnchor approach was the best DML for all datasets adopted in this work, achieving the $2^\circ$, $1^\circ$, and $2^\circ$ ranks in the AID, UCMerced and RESISC45 datasets, respectively. Finally, ResNet50 architecture reached the best average accuracy among all DLA adopted in this work for two datasets (UCMerced and RESISC45).

In the experiments with pre-trained CNNs, it is possible to observe that PT-CNN achieved to be better than the average accuracy (in black) of the DML approaches only when it uses ResNet18. However, when PT-CNN is compared to the best DML$+$DLA approaches in each dataset, PT-CNN did not surpass the performance of them. For instance, in the AID dataset, the best PT-CNN using ResNet18 achieved $85.29\%$ of accuracy against $87.37\%$ achieved by SupCon$+$ResNet18. In the UCMerced dataset, PT-CNN using ResNet18 achieved $69.68\%$ of accuracy against $77.70$ achieved by ProxyAnchor$+$VGG19. Finally, in the RESIC45 dataset, the best PT-CNN using ResNet18 achieved $89.75\%$ of accuracy against $91.60\%$ achieved by  NNGK$+$ResNet50.

\begin{table}[ht!]
    \centering
     \caption{Accuracy results  ($\%$), standard deviation ($\pm$), and ranking position (\textit{$@$R}) of six different deep metric learning approaches ($DML$) and four deep learning architectures ($DLA$) for three well-known aerial scene datasets using $5$-fold cross validation protocol. The best $DML+DLA$ tuple results for each dataset are in gray. The best DLA for each dataset are in black. PT-CNN means the Pre-Trained CNN as a traditional baseline from the literature.}
     \vspace{0.25cm}
\resizebox{14.5cm}{!}{
    \begin{tabular}{|c|ccccccccc|c|c|} \hline 
\multirow{2}{*}{\textbf{Datasets}} &\multirow{1}{*}{\textbf{DML}}  &\multicolumn{8}{c|}{\textbf{Deep Learning Architectures (DLA)}}  &\multirow{2}{*}{\textbf{Average}}&\multirow{1}{*}{\textbf{Final}}\\ \cline{3-10} 
 & \textbf{Approaches} & \textbf{ResNet18} & \textbf{\textit{$@$R}} & \textbf{ResNet50} &\textbf{\textit{$@$R}} &\textbf{VGG16} &\textbf{\textit{$@$R}} &\textbf{VGG19}&\textbf{\textit{$@$R}}&&\textbf{\textit{$@$R}}\\ \cline{1-12} 
\multirow{8}{*}{\textbf{AID}}
&Contrastive & 82.11 $\pm$ 1.51 &6$^{\circ}$& 76.91 $\pm$ 0.81 &6$^{\circ}$& 78.04 $\pm$ 2.01 &4$^{\circ}$& 61.56 $\pm$ 5.16 &5$^{\circ}$& 74.66 & 5$^{\circ}$\\
&NNGK & 82.23 $\pm$ 1.15 &5$^{\circ}$& 85.81 $\pm$ 0.74 &2$^{\circ}$& 77.19 $\pm$ 0.82 &5$^{\circ}$& 77.87 $\pm$ 0.85 &4$^{\circ}$& 80.78 & 4$^{\circ}$\\
&ProxyAnchor & 82.35 $\pm$ 0.95 &4$^{\circ}$& 82.11 $\pm$ 5.96 &5$^{\circ}$& 86.16 $\pm$ 0.64 &1$^{\circ}$& 86.10 $\pm$ 1.06 &1$^{\circ}$& 84.18 & 2$^{\circ}$\\
&SoftTriple & 85.40 $\pm$ 1.29 &2$^{\circ}$& 84.19 $\pm$ 1.50 &3$^{\circ}$& 85.48 $\pm$ 0.55 &2$^{\circ}$& 84.96 $\pm$ 0.86 &2$^{\circ}$& 85.01 & \best{\textbf{1$^{\circ}$}}\\
&SupCon & \best{87.37 $\pm$ 1.39} &1$^{\circ}$& 86.79 $\pm$ 2.13 &1$^{\circ}$& 65.29 $\pm$ 13.44 &6$^{\circ}$& 53.12 $\pm$ 17.74 &6$^{\circ}$& 73.14 & 6$^{\circ}$\\ 
&Triplet & 83.74 $\pm$ 0.61 &3$^{\circ}$& 83.07 $\pm$ 0.65 &4$^{\circ}$& 80.70 $\pm$ 1.69 &3$^{\circ}$& 78.99 $\pm$ 1.77 &3$^{\circ}$& 81.62 & 3$^{\circ}$\\  \cline{2-12} 
\cline{2-12}  & \textbf{Average} & \bestcnn{83.87} && 83.15 && 78.81 && 73.77& &\multirow{2}{*}{\textbf{Average}}& \textbf{Final}  \\  \cline{2-10}  
  
 & {PT-CNN} & 85.29 $\pm$ 1.82   && 78.70 $\pm$ 0.86   &&  80.27 $\pm$ 0.89   && 56.95 $\pm$ 7.95   &&  &\textbf{\textit{$@$R}}\\  \hline

\multirow{8}{*}{\textbf{UCMerced}} 
&Contrastive &   64.52 $\pm$ 5.08 &6$^{\circ}$&   60.24 $\pm$ 2.73 &6$^{\circ}$&   58.95 $\pm$ 14.88 &6$^{\circ}$&   63.37 $\pm$ 10.79 &4$^{\circ}$&   61.77 &   6$^{\circ}$\\
&NNGK &   70.64 $\pm$ 0.33 &2$^{\circ}$&   74.81 $\pm$ 1.40 &2$^{\circ}$&   59.41 $\pm$ 1.21 &5$^{\circ}$&   62.33 $\pm$ 2.86 &5$^{\circ}$&   66.80 &   4$^{\circ}$\\
&ProxyAnchor &   65.43 $\pm$ 4.88 &5$^{\circ}$&   71.02 $\pm$ 3.61 &3$^{\circ}$&   74.75 $\pm$ 3.80 &2$^{\circ}$&   \best{77.70 $\pm$ 4.10} &1$^{\circ}$&   72.22 &   \best{\textbf{1$^{\circ}$}}\\
&SoftTriple &   68.60 $\pm$ 2.85 &4$^{\circ}$&   66.51 $\pm$ 3.40 &5$^{\circ}$&   75.59 $\pm$ 2.57 &1$^{\circ}$&   73.73 $\pm$ 2.41 &2$^{\circ}$&   71.11 &   2$^{\circ}$\\
&SupCon &   68.63 $\pm$ 3.60 &3$^{\circ}$&   76.35 $\pm$ 3.46 &1$^{\circ}$&   66.90 $\pm$ 5.02 &4$^{\circ}$&   49.89 $\pm$ 21.69 &6$^{\circ}$&   65.44 &   5$^{\circ}$\\
&Triplet &   74.67 $\pm$ 3.19 &1$^{\circ}$&   66.70 $\pm$ 6.53 &4$^{\circ}$&   70.40 $\pm$ 3.43 &3$^{\circ}$&   70.94 $\pm$ 3.16 &3$^{\circ}$&   70.67 &   3$^{\circ}$\\ \cline{2-12} 
 &  \textbf{Average} & 68.75 && \bestcnn{69.27} && 67.67 && 66.32 &&\multirow{2}{*}{\textbf{Average}}&\textbf{Final}\\ 
 \cline{2-10}  
  & {PT-CNN} & 69.68 $\pm$	1.62   && 68.19 $\pm$ 1.82    &&  64.73 $\pm$ 2.67   && 63.75 $\pm$	2.23   &&  & \textbf{\textit{$@$R}} \\  \hline

\multirow{8}{*}{\textbf{RESISC45}} 
& Contrastive & 64.50 $\pm$  1.90 &3$^{\circ}$& 82.75 $\pm$  1.43 &6$^{\circ}$& 78.23 $\pm$  2.44 &6$^{\circ}$& 57.80 $\pm$  4.61 &6$^{\circ}$& 70.82 & 6$^{\circ}$\\
& NNGK & 89.55 $\pm$  0.28 &1$^{\circ}$& \best{91.60 $\pm$  0.18} &1$^{\circ}$& 87.19 $\pm$  0.31 &5$^{\circ}$& 87.90 $\pm$  0.37 &1$^{\circ}$& 89.06 & \best{\textbf{1$^{\circ}$}}\\
 & ProxyAnchor & 86.83 $\pm$  1.22 &2$^{\circ}$& 88.53 $\pm$  0.62 &5$^{\circ}$& 89.99 $\pm$  0.15 &1$^{\circ}$& 82.26 $\pm$  0.29 &5$^{\circ}$& 86.90 & 2$^{\circ}$\\
 & SoftTriple & 62.43 $\pm$  0.78 &5$^{\circ}$& 90.75 $\pm$  0.22 &3$^{\circ}$& 86.44 $\pm$  0.25 &3$^{\circ}$& 85.66 $\pm$  0.30 &3$^{\circ}$& 81.32 & 4$^{\circ}$\\
 & SupCon & 62.44 $\pm$  1.43 &4$^{\circ}$& 89.63 $\pm$  0.60 &4$^{\circ}$& 87.81 $\pm$  0.75 &2$^{\circ}$& 86.09 $\pm$  1.07 &2$^{\circ}$& 81.49 & 3$^{\circ}$\\
 & Triplet & 57.03 $\pm$  0.60 &6$^{\circ}$& 91.31 $\pm$  0.22 &2$^{\circ}$& 85.43 $\pm$  0.29 &4$^{\circ}$& 83.87 $\pm$  0.76 &4$^{\circ}$& 79.41 & 5$^{\circ}$\\
  \cline{2-12}

 & \textbf{Average}  & 70.46 && \bestcnn{89.09} && 85.85 && 80.6 &&\multicolumn{1}{c}{} &\multicolumn{1}{c}{} \\  \cline{2-10}  
 
 & {PT-CNN} & 89.75 $\pm$	0.30   && 82.98 $\pm$	0.10   &&  83.50 $\pm$ 0.88   && 83.23 $\pm$ 1.49   &&\multicolumn{1}{c}{}  & \multicolumn{1}{c}{}\\ \cline{1-10}
\end{tabular}
}
\label{tab:overall}
\end{table}
%\faf{Adicionar resultados de CNNs pre-treinadas.}

\subsection{Computing Diversity among Classifiers}

\begin{figure*}[ht!]
    \centering
\includegraphics[scale=.2]{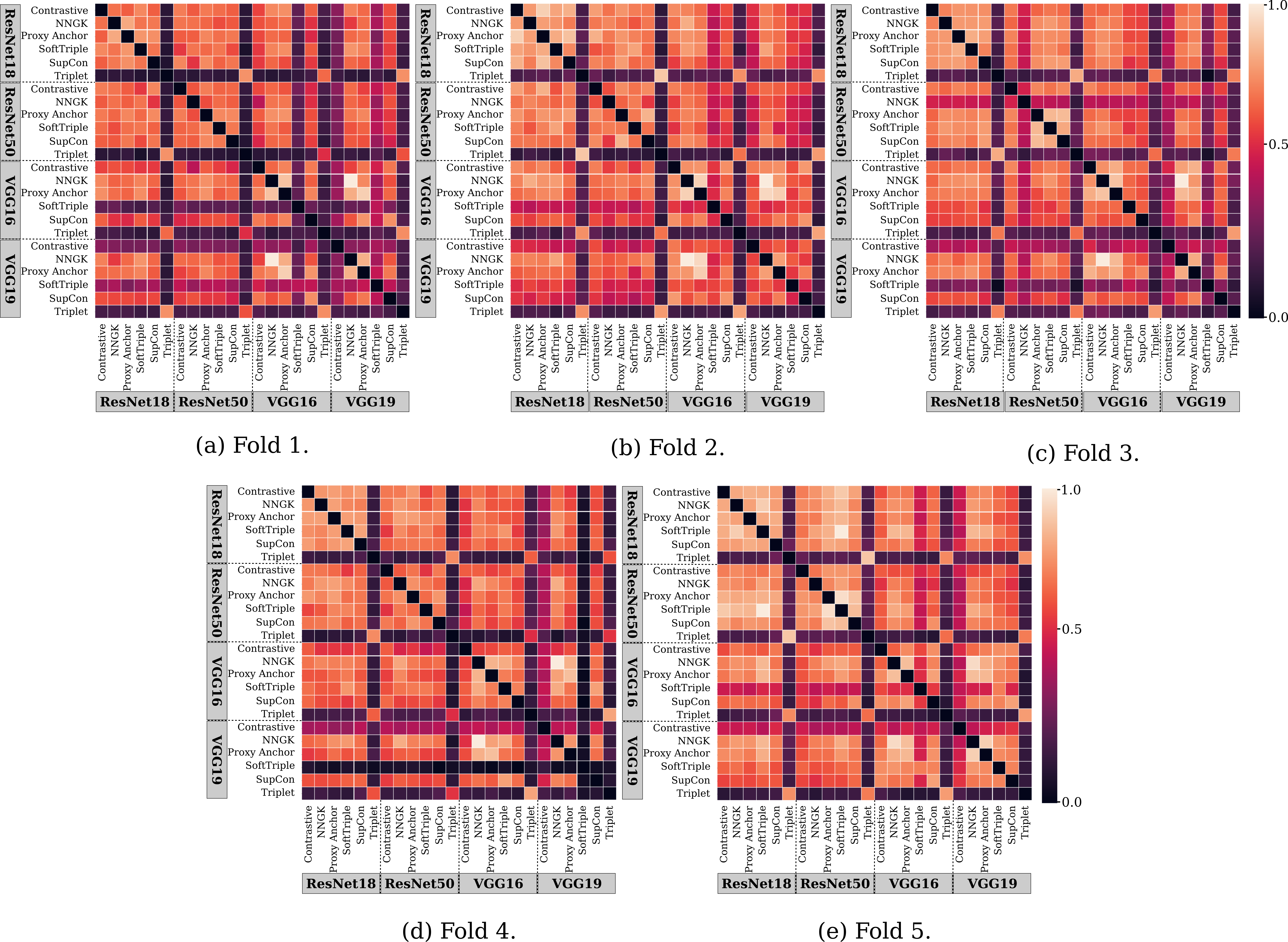}
\caption{Correlation Coefficient ($\rho$) as a diversity measurement computed among $24$ classifiers (DML$+$DLA) in a 5-folds cross validation protocol using AID dataset. A similar behavior occurred as using the UCMerced and RESISC45 datasets.}
\label{fig:div}
\end{figure*}

Now, in these experiments, the objective is to check for possible complementary information among all available classifiers created by combined tuples (DML$+$DLA). Therefore, we adopted the Correlation Coefficient measure to compute the diversity existing among the pool of classifiers. Figure~\ref{fig:div} shows the diversity measure scores (relationship matrices) for twenty-four different classifiers (DML$+$DLA) available in this work using the AID dataset. However,  similar behavior could be observed when using the UCMerced and RESISC45 datasets.

As mentioned earlier, diversity is greater if the $\rho$ correlation coefficient score is smaller between a pair of classifiers. Therefore, darker colors (closer to the zero score) indicate pairs of classifiers that may be better suited to be combined into an ensemble of classifiers. The lighter colors (closer to a score) are pairs of classifiers that have similar behavior in the validation set and do not carry additional information in a possible combination between them.

In Figure~\ref{fig:div}, it is possible to observe that the VGG-based classifiers  are less correlated with the ResNet ones, thus they can be present together in an ensemble of classifiers. Furthermore, triplet approaches with different CNN architectures appear to be combinable with all other classifiers present in this work. Therefore, as there is complementary information among some of the available classifiers, we can improve the classification results in the target task by employing classifier fusion strategies.

\subsection{Creating Ensembles of Classifiers}

To improve classification results in the scene classification task, we employ two classifier fusion strategies for combining the most suitable classifiers composed of the tuple (DML$+$DLA) into ensembles of classifiers.

First, we adopted majority voting (MV) due to its simplicity. MV predicts the final sample class through the majority voting of
all classifiers in the pool (a total of $24$ classifiers or tuples composed of DML$+$DLA).  Another strategy adopted in this experiment is the Univariate Marginal Distribution Algorithm (UMDA), which it is known as one of the simplest Estimation of Distribution Algorithms (EDAs). In this experiment, we used the Meta-heuristic Algorithms Python Library toolkit~\footnote{\tiny{\url{https://github.com/gugarosa/evolutionary_ensembles}} (As of February 2023)} to select the most suitable classifiers from the pool by UMDA strategy.

\begin{figure}[ht!]
\centering
\begin{tabular}{cc}
  \includegraphics[scale=.42]{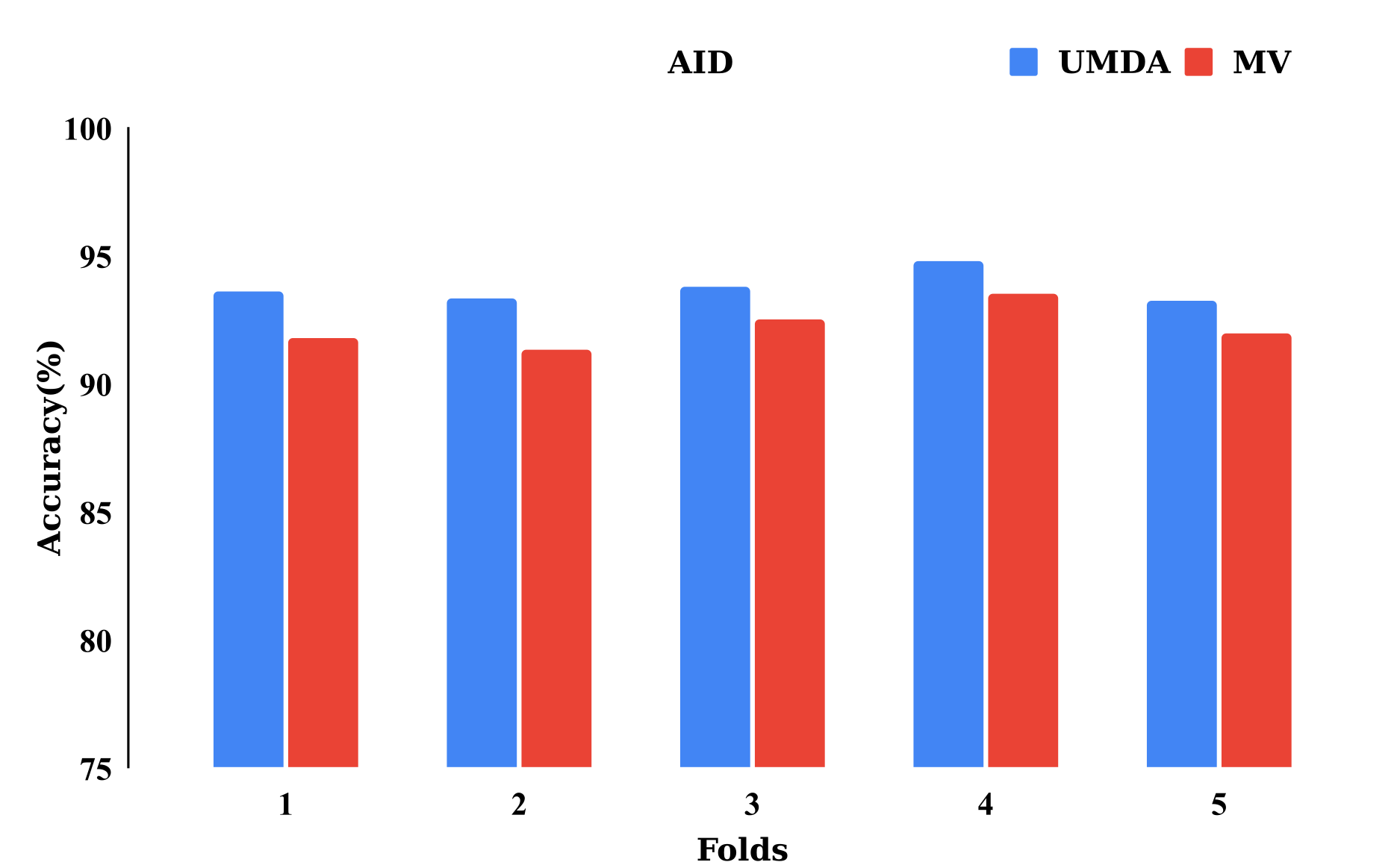}   & \includegraphics[scale=.42]{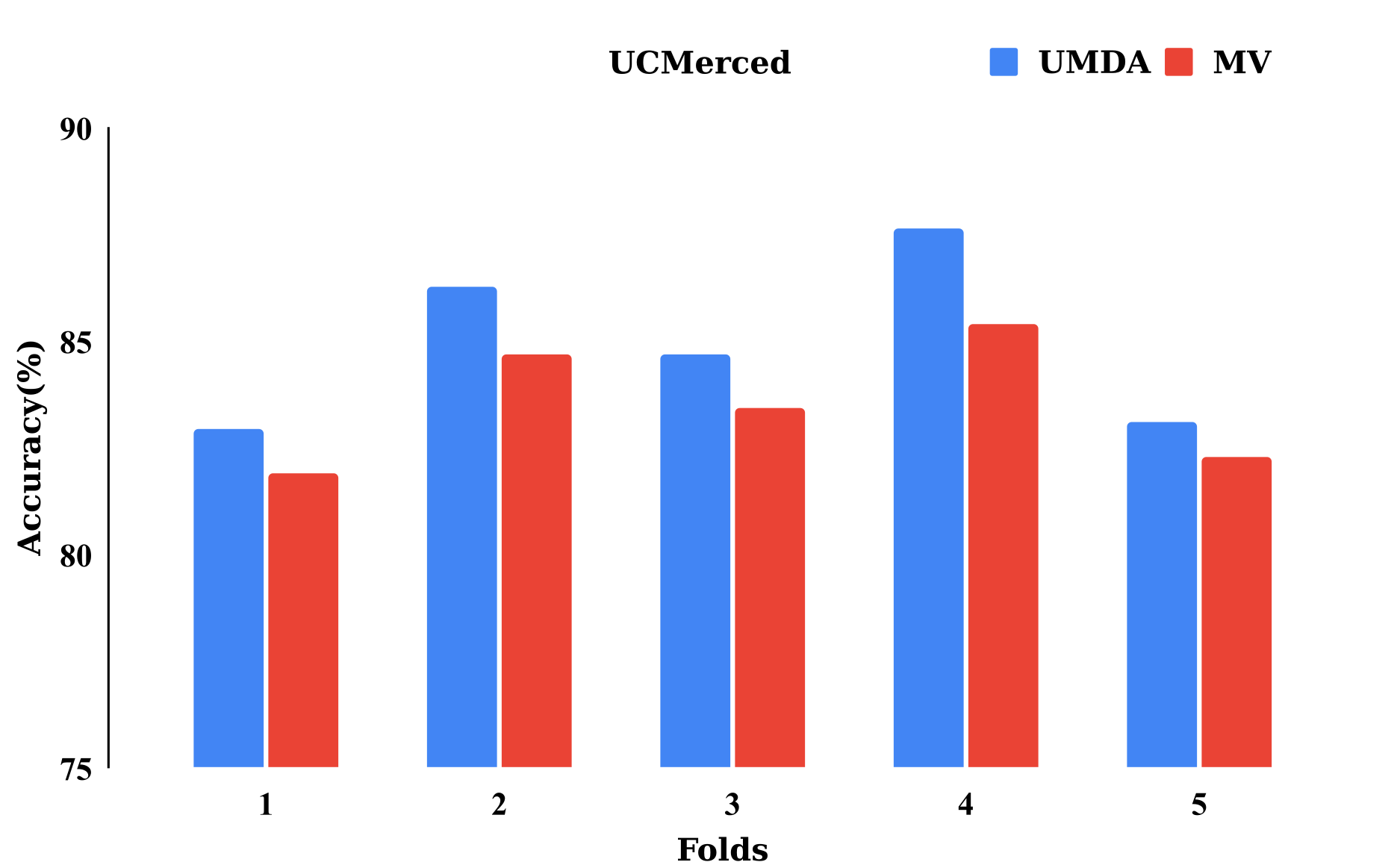} \\
  (a) AID. & (b) UCMerced. \\
   \multicolumn{2}{c}{\includegraphics[scale=.42]{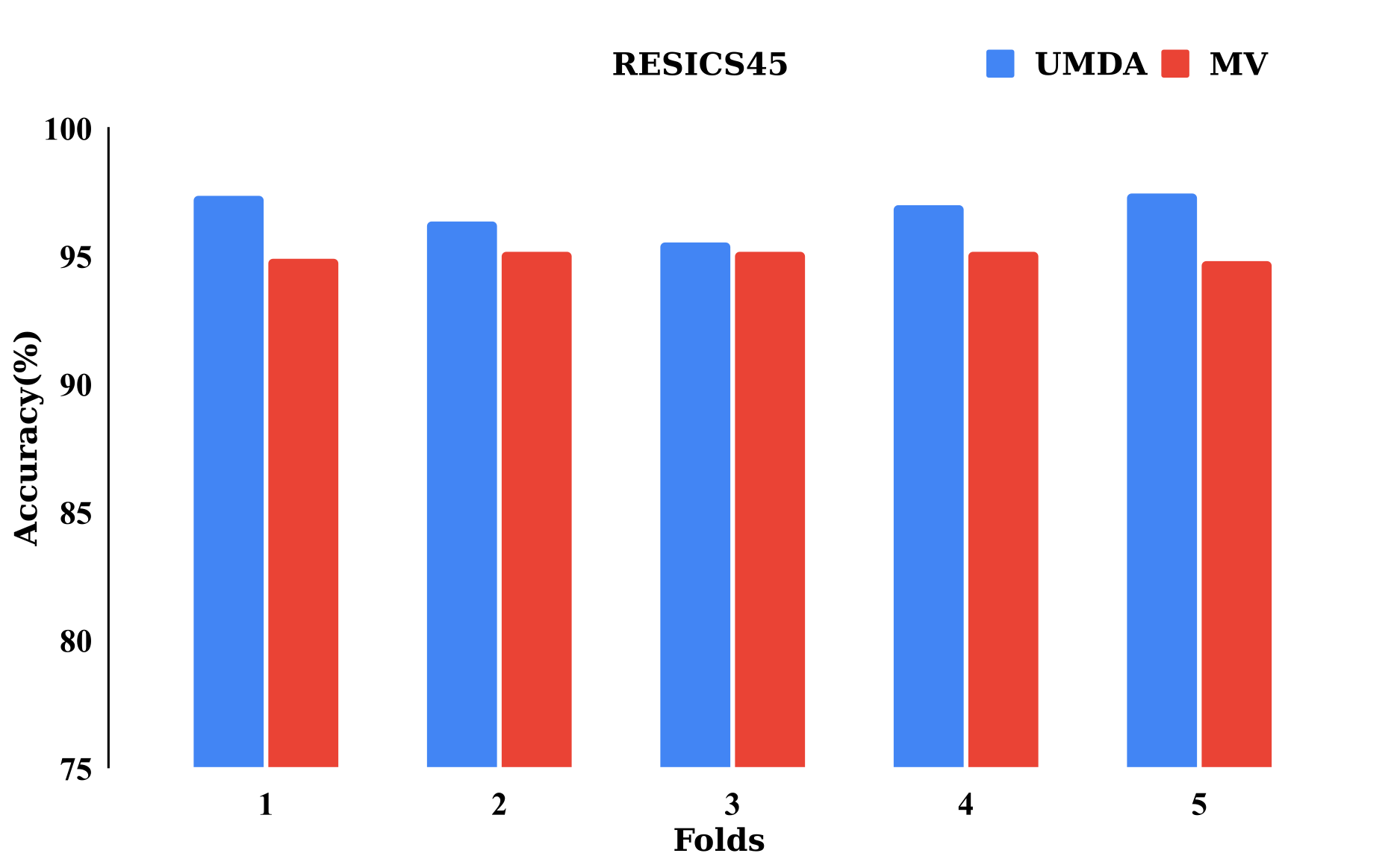}}\\
   \multicolumn{2}{c}{(c) RESISC45. }   
\end{tabular}
\caption{UMDA and MV classification results (\%) for each of the five folds in the three adopted aerial scenes datasets (AID, UCMerced, and RESISC45).}
\label{fig:ensemble_res}
\end{figure}

Figures~\ref{fig:ensemble_res} shows the classification results of the two fusion strategies (MV and UMDA) for AID, UCMerced, and RESISC45 datasets in a 5-folds cross-validation protocol. It is possible to observe that UMDA (in blue) achieved better results than MV (in red) for all five folds in the three different aerial scene datasets.

Table~\ref{tab:ensemble} shows the classification results and standard deviation of the two classifier fusion strategies. We also present the number of classifiers used by each strategy to predict the samples from the test set. Furthermore, we also included the best tuple (DML$+$DLA) from the previous experiments for comparison purposes.

As it is possible to observe that the UMDA strategy achieved slightly better results than MV for all three datasets, using almost 50\% less of the available classifiers ($[.]$) to predict the final sample classes. It presents a good classification compared to the best tuple (DML$+$DLA) with relative gains of $7.32\%$, $9.29\%$, and $5.60\%$ in the AID, UCMerced, and RESISC45 datasets, respectively. Finally, UMDA strategy also achieved better classification results when compared to the best pre-trained CNN (PT-CNN) with relative gains of $9.94\%$, $21.87\%$, and $7.77\%$ in the same datasets.

\begin{table}[ht!]
    \centering
     \caption{Accuracy results (\%), standard deviation ($\pm$), and the number of classifiers ($[.]$) used by fusion strategies MV and UMDA in a 5-folds cross-validation protocol. With \textbf{$^*$} are best baseline classifiers for each dataset.} \vspace{0.25cm}
   \resizebox{11cm}{!}{  
    \begin{tabular}{|c|ccc|} \hline
        \textbf{Fusion} & \multicolumn{3}{c|}{\textbf{Datasets}}  \\ \cline{2-4}
        \textbf{Strategies} & \textbf{AID} & \textbf{UCMerced} & \textbf{RESISC45} \\ \hline
        MV&92.25$\pm$0.85 $[24]$ & 83.54$\pm$1.50 $[24]$ & 95.03$\pm$0.20 $[24]$\\ 
        UMDA & \textbf{93.77}$\pm$0.63 $[11]$ & \textbf{84.92}$\pm$2.03 $[14]$ & \textbf{96.73}$\pm$0.80 $[11]$\\\hline
         \multicolumn{4}{|c|}{\textbf{Best Baseline Approaches}} \\ \hline
        {DML$+$DLA}\textbf{$^*$} & 87.37 $\pm$ 1.39 & 77.70 $\pm$ 4.10  & 91.60 $\pm$ 0.18\\ \hline
        {PT-CNN}\textbf{$^*$} & 85.29 $\pm$  1.82  &69.68  $\pm$ 1.62  &  89.75 $\pm$ 0.30\\ \hline
        \hline
        \multicolumn{4}{|c|}{\textbf{Relative Gain of the UMDA $\times$ Best approaches}} \\ \hline
        {DML+DLA}\textbf{$^*$} & \textbf{7.32}  & \textbf{9.29}  & \textbf{5.60}  \\ \hline
        {PT-CNN}\textbf{$^*$} & \textbf{9.94}  & \textbf{21.87}  & \textbf{7.77}  \\ \hline        
    \end{tabular}
    }
    \label{tab:ensemble}
\end{table}

\section{Conclusion}

Remote sensing aerial scene classification is a challenging task due to a large intra-class variability existing in the datasets from the literature. Similar to the field of computer vision, using convolutional neural networks (CNN) to perform a traditional classification task also yields surprising results in this application. However, another way less used in remote sensing to perform such task is through the use of deep metric learning (DML) approaches. For this reason, in this work, we adopted six different DML approaches (Contrastive, ProxyAnchor, SoftTriple, SupCon, Triplet and NNGK) and four well-known deep learning architectures (DLA -- ResNet18, ResNet50, VGG16 and VGG19), for aerial scene classification task. 

As findings of this work, we could show through experiments that:

\begin{enumerate} 
   \item DML approaches have shown to be highly dependent on DLA, achieving classification results with large variance on the same dataset;
   \item  DML approaches can achieve better results than pre-trained architectures (PT-CNN) in all datasets used in this work;
   \item The ResNet50 as DLA in the tuple DML$+$DLA achieved better average accuracies in two out of three datasets;
   \item In general, ProxyAnchor approach was the best among the six DML approaches compared in this work;
   \item The DML-based classifiers (DML$+$DLA) proved to provide complementary information among them, allowing their combination
 and the creation of a strongerPlease provide a single file containing your manuscript now. Data included in your manuscript may be used to populate information for you later in the submission process. final classifier (ensemble of classifiers);
   \item  The UMDA strategy proved to be better than the simple majority voting (MV) for all datasets using almost 50\% of the available classifiers for the construction of the final ensemble of classifiers;
   \item  Finally, the ensembles of classifiers created by UMDA achieved at least $5.6\%$ and $7.77\%$, in relative gains over the best tuple DML$+$DLA and PT-CNN, respectively.
   
\end{enumerate}

\section*{Acknowledgement}
The authors are grateful to the S{\~a}o Paulo Research Foundation (FAPESP) grants 2017/25908-6, 2018/23908-1, and 2021/01870-5, to the National Council for Scientific and Technological Development (CNPq), to the National Laboratory for Scientific Computing (LNCC/MCTI, Brazil) for providing HPC resources of the SDumont supercomputer, which have contributed to the research results reported within this paper\footnote{SDumont - \url{http://sdumont.lncc.br}}.




\bibliographystyle{elsarticle-num}
\bibliography{ref}

\end{document}